\documentclass[conference]{IEEEtran}
\IEEEoverridecommandlockouts
\usepackage{cite}
\usepackage{amsmath,amssymb,amsfonts}
\usepackage{algorithmic}
\usepackage{graphicx}
\usepackage{textcomp}
\usepackage{xcolor}

\usepackage{amsmath,graphicx}
\usepackage{booktabs}
\usepackage{tabularx}
\usepackage{multirow}
\usepackage{mathtools}
\usepackage{arydshln}
\usepackage{gensymb}
\usepackage{textcomp}
\usepackage{amsfonts}
\usepackage{amsthm}
\usepackage{eucal}
\usepackage{microtype}
\usepackage{hyperref}       % hyperlinks
\usepackage{url}            % simple URL typesetting
\usepackage{nicefrac}       % compact symbols for 1/2, etc.
\usepackage{subfigure}
\usepackage{xcolor}
\usepackage{orcidlink}

\def\BibTeX{{\rm B\kern-.05em{\sc i\kern-.025em b}\kern-.08em
    T\kern-.1667em\lower.7ex\hbox{E}\kern-.125emX}}
\begin{document}

\title{Transfer Learning for Keypoint Detection in Low-Resolution Thermal TUG Test Images}

\author{
    \IEEEauthorblockN{
        Wei-Lun Chen\IEEEauthorrefmark{1}\IEEEauthorrefmark{2}, 
        Chia-Yeh Hsieh\IEEEauthorrefmark{3}, 
        Yu-Hsiang Kao\IEEEauthorrefmark{4}, 
        Kai-Chun Liu\IEEEauthorrefmark{5}, 
        Sheng-Yu Peng\IEEEauthorrefmark{6},
        and Yu Tsao\IEEEauthorrefmark{1}
    }
    
    \IEEEauthorblockA{
        \IEEEauthorrefmark{1}Research Center for Information Technology Innovation, Academic Sinica, Taiwan
    }
    \IEEEauthorblockA{
        \IEEEauthorrefmark{2}Graduate Institute of Electrical Engineering, National Taiwan University, Taiwan
    }
    \IEEEauthorblockA{
        \IEEEauthorrefmark{3}Bachelor’s Program in Medical Informatics and Innovative Applications, Fu Jen Catholic University, Taiwan
    }
    \IEEEauthorblockA{
        \IEEEauthorrefmark{4}Department of Electrical Engineering, National Taiwan University, Taiwan
    }
    \IEEEauthorblockA{
        \IEEEauthorrefmark{5}College of Information and Computer Sciences, University of Massachusetts Amherst, MA, 01003, USA
    }
    \IEEEauthorblockA{
        \IEEEauthorrefmark{6}Department of Electrical Engineering, National Taiwan University of Science of Technology, Taiwan
    }
}

\maketitle

\begin{abstract}
This study presents a novel approach to human keypoint detection in low-resolution thermal images using transfer learning techniques. We introduce the first application of the Timed Up and Go (TUG) test in thermal image computer vision, establishing a new paradigm for mobility assessment. Our method leverages a MobileNetV3-Small encoder and a ViTPose decoder, trained using a composite loss function that balances latent representation alignment and heatmap accuracy. The model was evaluated using the Object Keypoint Similarity (OKS) metric from the COCO Keypoint Detection Challenge. The proposed model achieves better performance with AP, AP50, and AP75 scores of 0.861, 0.942, and 0.887 respectively, outperforming traditional supervised learning approaches like Mask R-CNN and ViTPose-Base. Moreover, our model demonstrates superior computational efficiency in terms of parameter count and FLOPS. This research lays a solid foundation for future clinical applications of thermal imaging in mobility assessment and rehabilitation monitoring.
\end{abstract}


\section{Introduction}
\begin{figure}[htbp]
\centerline{\includegraphics[width=0.73\columnwidth]{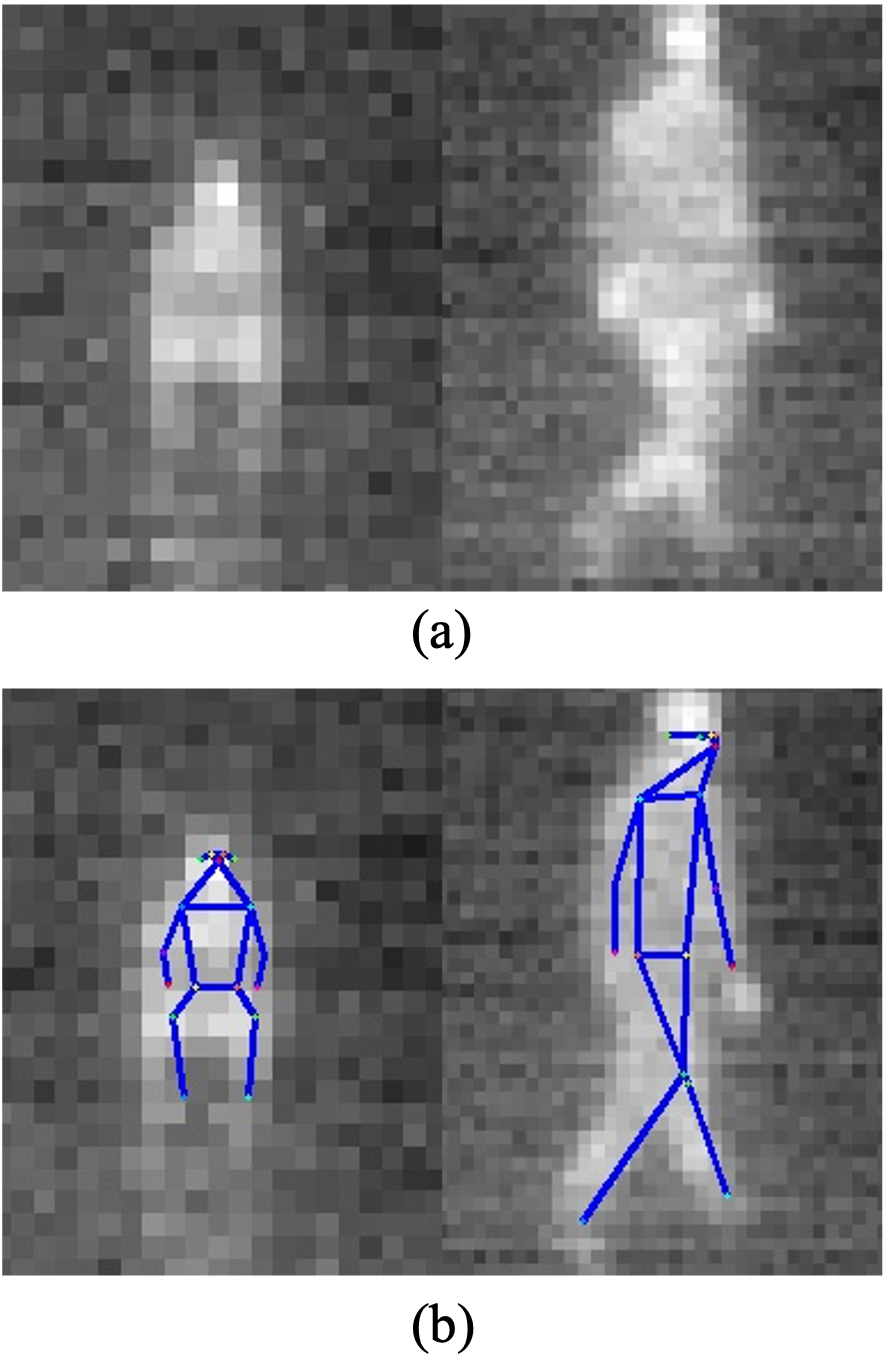}}
\caption{(a) The original thermal image. (b) The thermal image with annotated human keypoints which generated from proposed model.}
\label{fig:compare}
\end{figure}

Low-resolution thermal imaging cameras offer a cost-effective and privacy-compliant solution to monitor patient activities in hospitals. However, the original low-resolution thermal images are challenging to interpret for untrained caregivers. By incorporating appropriate thermal image annotations, such as human keypoints and skeletal structures, the comprehensibility of original low-resolution thermal images can be significantly enhanced, as shown in Fig. \ref{fig:compare}. Furthermore, human keypoints can provide additional clinically relevant information to healthcare professionals. For instance, the analysis of keypoints from various regions of the body can provide insight into gait balance \cite{nuckolsMechanicsWalkingRunning2020, stenumClinicalGaitAnalysis2024}. Additionally, previous studies indicated that incorporating keypoint information into Human Activity Recognition (HAR) systems can significantly enhance recognition performance. \cite{batchuluunActionRecognitionThermal2019, batchuluunActionRecognitionThermal2021}.

To provide accurate annotations, a reliable human keypoint detection model is essential. However, training such a model from scratch presents significant challenges. This process requires a substantial volume of annotated data \cite{xuViTPoseSimpleVision2022, xuViTPoseVisionTransformer2023}, while the task of annotating low-resolution thermal images is arduous and labor-intensive. Moreover, the training procedure demands considerable computational resources, further complicating the development of an effective keypoint detection model.

Transfer learning techniques offer an effective approach to address these challenges \cite{zhuangComprehensiveSurveyTransfer2020, bommasaniOpportunitiesRisksFoundation2022, dhekaneTransferLearningHuman2024}. By leveraging mature generative models developed for RGB computer vision, it becomes feasible to train models for human keypoint detection in low-resolution thermal images using a limited dataset, without the need for additional manual annotation. This approach not only mitigates the labeling burden but also capitalizes on existing knowledge in the field. It is worth noting that transfer learning yields the additional benefit of model compression \cite{zhuangComprehensiveSurveyTransfer2020, houlsbyParameterEfficientTransferLearning2019}. This reduction in model size enhances the potential for future applications in edge AI, where computational resources are often constrained. The smaller footprint of these models makes them more suitable for deployment in resource-limited environments, thereby expanding the scope of their practical implementation.

This study utilizes low-resolution thermal images and RGB images of the 3-meter Timed Up and Go (TUG) test as training samples for the model. The TUG test encompasses movements at various distances, in different moving directions, and includes both static and dynamic actions. The TUG dataset allows for the validation of the robustness of the proposed model in keypoint detection across diverse activity types and distance conditions. By incorporating this comprehensive set of movements, the study aims to assess the model's performance in accurately identifying human keypoints under a range of scenarios that mimic real-world applications. In summary, the main contributions of this paper are as follows:
\begin{itemize}
\item This study presents the first model designed specifically for human keypoint detection in low-resolution thermal images.
\item This study leverages transfer learning techniques to significantly reduce the costs associated with data annotation and model training, while simultaneously achieving a reduction in model size.
\item This study represents the first application of TUG test in the field of thermal image computer vision, validating its feasibility in this context. This work lays the foundation for future clinical research in this domain.
\end{itemize}

\begin{figure*}[ht]
\centerline{\includegraphics[width=1.85\columnwidth]{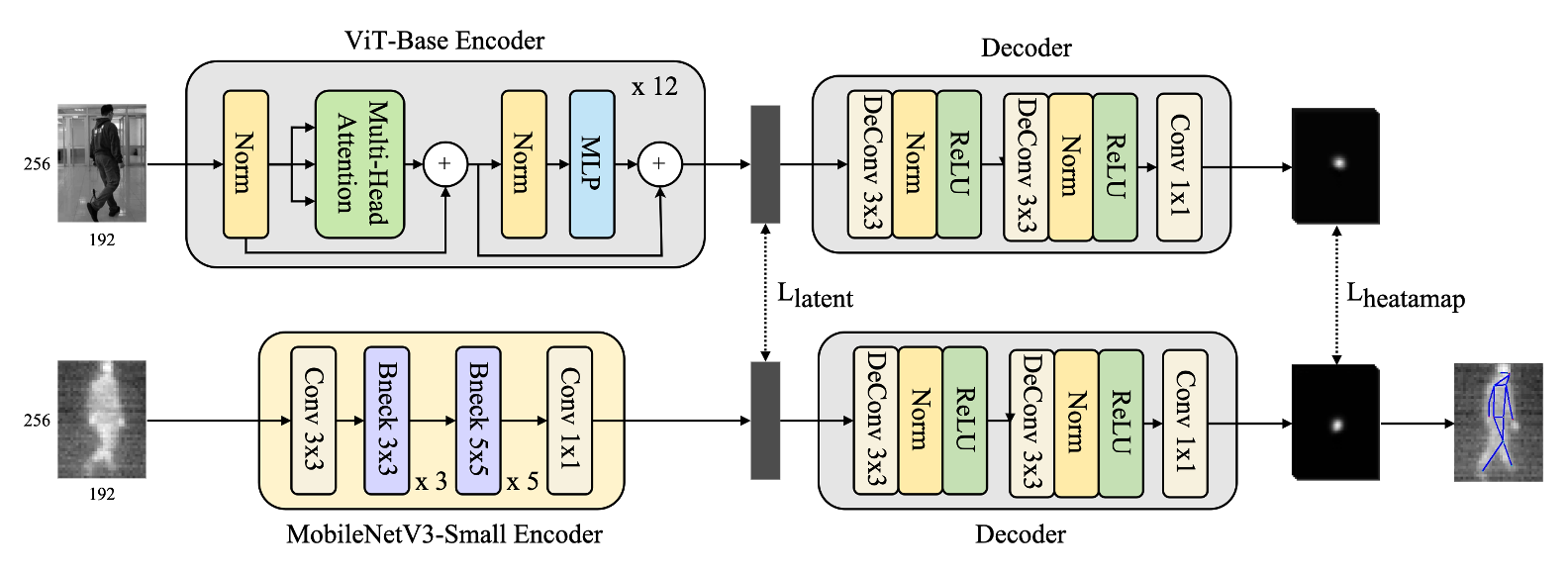}}
\caption{The architecture for keypoint detection via transfer learning. The blocks with a gray background indicate pre-trained components that are frozen during training, while the blocks with a yellow background represent the unfrozen parts that are fine-tuned. The input data consists of human bounding box detections that have been cropped and resized.}
\label{fig:model}
\end{figure*}

\section{Methodology}
The experimental workflow of this study can be broadly divided into two key steps: human bounding box detection and keypoint detection.

\subsection{Human Bounding Box Detection}
The bounding box detection component serves a crucial role in identifying the location of the human subject within the thermal image. This preprocessing step ensures that the subsequent keypoint detection model is provided with input data containing a person. Moreover, compared to directly inputting the entire thermal image, preprocessing with human bounding box detection ensures a higher proportion of the human subject in the frame.

In this study, the bounding box detection is performed using a Faster R-CNN \cite{girshickFastRCNN2015, renFasterRCNNRealTime2016} model with a ResNet-50 \cite{heDeepResidualLearning2015} backbone, trained for human recognition and localization within the thermal images. Upon successful detection, the center of the bounding box is used to extract and resize both the thermal and RGB images to a consistent size of $256 \times 192$ pixels.

\subsection{Keypoint Detection}
As shown in Fig. \ref{fig:model}. After the resizing and cropping steps, the extracted thermal images and RGB images are used for the transfer learning-based training of the keypoint detection model. For RGB images, a pre-trained ViTPose-Base \cite{xuViTPoseSimpleVision2022, xuViTPoseVisionTransformer2023} model with a classic decoder is employed as the teacher model. 

For the thermal image model, the MobileNetV3-Small \cite{howardSearchingMobileNetV32019} network layers are used as the encoder, coupled with the same ViTPose classic decoder as the RGB teacher model, forming the overall student model architecture. The encoder details comprise 1 layer of $3 \times 3$ kernel convolutional layer followed by 8 bottleneck layers. Finally, a $1 \times 1$ kernel convolutional layer is employed to adjust the latent representation to the desired length.

The training objective is to leverage transfer learning techniques to extract suitable latent representations from the thermal images, and then utilize the pre-trained decoder to obtain the keypoint heatmap $K$. The heatmap $K$ has a size of $64 \times 48 \times N_k$, where $64 \times 48$ represents the heatmap dimensions, and $N_k$ denotes the number of keypoints. In this study, the same 17 keypoint locations as the COCO dataset are used ($N_k = 17$). 

Ultimately, the information from the generated heatmaps is used to annotate the keypoint locations on the corresponding thermal images. This combined approach of transfer learning and shared decoder architecture allows the thermal image model to benefit from the knowledge distilled in the RGB teacher model, while preserving the ability to accurately localize the human keypoints within the low-resolution thermal imagery.

\subsection{Loss Function}
This study employs a composite loss function designed to optimize the two-step output of the model. The first component of the loss function utilizes L1 Loss for the latent representation, while the second component applies Adaptive Wing Loss \cite{wangAdaptiveWingLoss2019} to the keypoint heatmap. The relative contributions of these two loss components are balanced using a weighting factor $\beta$.

\begin{equation}
L_{total}= \beta L_{latent} + (1-\beta) L_{heatmap}\label{eq:loss}
\end{equation}

where $L_{latent}$ represents the L1 Loss applied to the latent space, and $L_{heatmap}$ denotes the Adaptive Wing Loss used for the keypoint heatmap. The hyperparameter $\beta \in [0,1]$ governs the trade-off between these two loss components, enabling precise fine-tuning of the model’s focus during the training process.
This formulation of the loss function enables the model to simultaneously optimize for both the quality of the latent representation alignment and the accuracy of the keypoint localization, thus enhancing the overall performance of the human pose estimation system in thermal imagery.

\section{Experiment}

\subsection{Data preparation}
This study utilizes an embedded thermal imaging camera (MIR8060B1) and an RGB camera to collect data from 3-meter Timed Up and Go (TUG) tests for model training. The thermal images have a resolution of $80 \times 60$ pixels, while the RGB images are captured at $1024 \times 576$ pixels. Both cameras operate at a frame rate of 8 fps. Data collection is synchronized using a multi-threaded approach, with approximately 1\% of asynchronous data manually discarded in post-processing.

Furthermore, both the thermal and RGB data undergo calibration and alignment procedures prior to use. This preprocessing step ensures spatial correspondence between the two imaging modalities, which is crucial for the subsequent transfer learning and model training processes.

This study collected data from 10 individuals with normal mobility, comprising 8 males and 2 females. Each participant performed the TUG test five times. To validate the model's efficacy and generalizability, the data was divided into 10 subsets using a leave-one-subject-out cross-validation approach.

\subsection{Implementation Detail}
During the transfer learning process for the keypoint detection model, only the weights of the MobileNetV3-Small encoder are updated through training. All other network components retain their pre-trained weights and remain frozen.

In this study, the weight parameter for the composite loss function \eqref{eq:loss} is set to $\beta = 0.4$. The model is optimized using the Adam optimizer with a learning rate of 0.01 and no weight decay. The training process employs a batch size of 50 and runs for 500 epochs. Additionally, a ReduceLROnPlateau scheduler \cite{mukherjeeSimpleDynamicLearning2019} is implemented to dynamically adjust learning parameters, and early stopping is employed to mitigate overfitting.

\subsection{Evalutaion Metrics}
This study employs the Object Keypoint Similarity (OKS) metric defined in the COCO Keypoint Detection Challenge, as the primary evaluation criterion. Based on this metric, Average Precision (AP) scores are calculated at various thresholds, specifically AP, AP50, and AP75.

\begin{equation}
\text{OKS} = \frac{\sum_i \exp\left(-\frac{d_i^2}{2s^2k_i^2}\right) \cdot \delta(v_i > 0)}{\sum_i \delta(v_i > 0)}
\label{eq:oks}
\end{equation}

\vspace{0.1cm}
\section{Results}
\subsection{Keypoint Detection Evaluation}

Our study presents a novel approach to keypoint detection in low-resolution thermal images using transfer learning techniques. The comparative analysis, detailed in Tables \ref{t:combined}, reveals the substantial advantages of our method over traditional supervised learning approaches. The proposed model achieves remarkable performance metrics, with an Average Precision (AP) of 0.861, AP50 of 0.942, and AP75 of 0.887. These results significantly outperform established models like Mask R-CNN \cite{heMaskRCNN2018} and ViTPose \cite{xuViTPoseSimpleVision2022}, demonstrating the efficacy of our transfer learning strategy in accurately localizing keypoints in challenging low-resolution thermal images.

A comparison between ViTPose and our transfer learning approach using a ViT-Base encoder reveals that transfer learning indeed enhances the model's suitability for low-resolution thermal image data. Furthermore, our investigation extended to non-attention architecture models as encoders. Notably, while ResNet-50 \cite{heDeepResidualLearning2015} as an encoder showed marginal improvements in AP and AP75 compared to our MobileNetV3-based model, the latter's parameter count is merely 32\% of ResNet-50's, and its floating-point operations per second (FLOPS) are only 63\% of ResNet-50's. This significant reduction in model complexity and computational requirements makes MobileNetV3 particularly advantageous for deployment in resource-constrained environments such as hospitals and edge AI applications, where computational efficiency is paramount.

\begin{table*}[!htb]
\caption{Comparison of Average Precision AP, AP50, AP75, number of parameters (Param), and Floating Point Operations per second (FLOPs) for different models in keypoint detection tasks.}
\label{t:combined}
\begin{center}
\begin{tabular}{llccccc}
\toprule[0.4mm]
Model   & Encoder   & AP    & AP50  & AP75  & Param (M) & FLOPs (G) \\
\midrule
Mask R-CNN \cite{heMaskRCNN2018} & ResNeXt-101 & $0.214 \pm 0.094$ & $0.484 \pm 0.157$ & $0.166 \pm 0.105$ & 122.217 & 526.722 \\
ViTPose \cite{xuViTPoseVisionTransformer2023} & ViT-Base & $0.043 \pm 0.010$ & $0.082 \pm 0.019$ & $0.038 \pm 0.009$ & 89.994 & 22.766 \\
\midrule
Ours & ViT-Base & $0.124 \pm 0.039$ & $0.228 \pm 0.065$ & $0.113 \pm 0.038$ & 89.994 & 22.766 \\
Ours & ResNet-50 & \underline{0.882 $\pm$ 0.088} & $0.938 \pm 0.055$ & \underline{0.897 $\pm$ 0.088} & 13.530 & 9.066 \\
\textbf{Ours} & \textbf{MobileNetV3} & \textbf{0.861 $\pm$ 0.071} & \underline{\textbf{0.942 $\pm$ 0.039}} & \textbf{0.887 $\pm$ 0.074} & \underline{\textbf{4.398}} & \underline{\textbf{5.696}} \\
\bottomrule
\multicolumn{7}{l}{* Bold indicates the purposed model, and underline indicates the best result.}
\end{tabular}%

\end{center}
\end{table*}

\subsection{Loss Weight Selection}

\begin{figure}[htbp]
\centerline{\includegraphics[width=0.85\columnwidth]{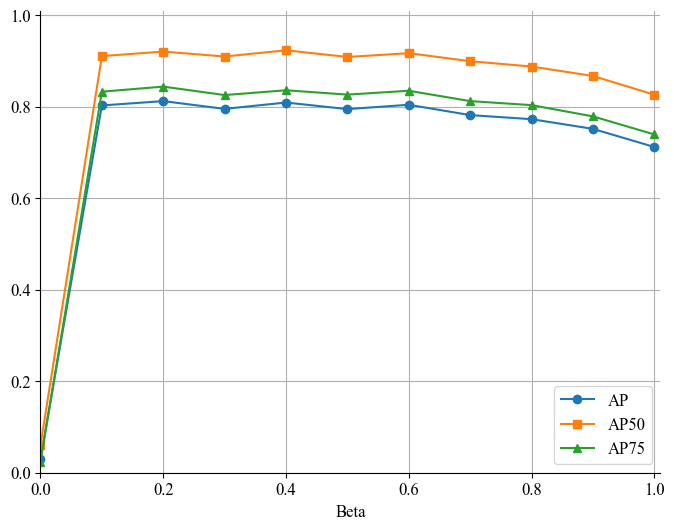}}
\caption{The effect of different $\beta$ values on the Average Precision (AP), AP50, and AP75 in keypoint detection tasks.}
\label{fig:beta}
\end{figure}

The impact of the weighting parameter $\beta$ in our composite loss function was systematically investigated, with results illustrated in Fig \ref{fig:beta}. A notable finding is the substantial improvement in precision as $\beta$ increases from 0.0 to 0.1, emphasizing the critical importance of incorporating latent representation in the loss function.

This observation carries significant implications for model design in thermal image keypoint detection. Training solely based on the final heatmap output ($\beta = 0.0$) proves suboptimal, suggesting that this approach fails to capture the full complexity of the task. In contrast, focusing exclusively on the latent representation alignment ($\beta = 1.0$) yields an AP exceeding 0.8, demonstrating the importance of learning appropriate intermediate representations for thermal imagery.
The most intriguing results emerge from a balanced approach. As $\beta$ varies from 0.1 to 0.9, we observe peak performance, indicating that a combination of latent representation and heatmap-based learning is crucial for optimal keypoint detection in thermal images. This synergy likely allows the model to leverage both high-level abstract features and fine-grained spatial information effectively.

Based on a comprehensive analysis of these results, we selected $\beta = 0.4$ as the optimal parameter for our model. This choice strikes a balance between latent representation learning and heatmap refinement, enabling the model to capture the nuances of low-resolution thermal images while maintaining precise keypoint localization.

\section{Conclusion}

This study marks a significant advancement in the field of human pose estimation using low-resolution thermal images. By successfully applying transfer learning techniques to the challenging domain of thermal image keypoint detection, we have demonstrated the potential for more efficient and accurate mobility assessments in clinical settings.

Our novel approach, which combines a MobileNetV3-Small encoder with a ViTPose decoder, achieves superior performance compared to traditional supervised learning methods. The model's high accuracy across various OKS thresholds (AP: 0.861, AP50: 0.942, AP75: 0.887) underscores its robustness and reliability in keypoint detection tasks.

The investigation into the composite loss function revealed the critical importance of balancing latent representation learning with heatmap accuracy. The optimal weight of $\beta = 0.4$ provides valuable insights for future research in transfer learning for thermal image analysis.

Furthermore, this study introduces the TUG test to the realm of thermal image computer vision, opening new avenues for non-invasive, privacy-preserving mobility assessments. The successful implementation of this widely used clinical test in a computer vision context bridges the gap between traditional clinical assessments and advanced technological solutions.

The computational efficiency of our model, evidenced by reduced parameter count and FLOPS, makes it particularly suitable for real-world applications where resources may be constrained. This efficiency, combined with high accuracy, positions our approach as a promising tool for various fields including healthcare, rehabilitation, and human-computer interaction.

In conclusion, this research not only advances the state-of-the-art in low-resolution thermal image keypoint detection but also lays a solid foundation for future clinical research. The integration of standardized clinical assessments with cutting-edge computer vision techniques offers exciting possibilities for enhancing the objectivity and granularity of mobility evaluations. The large-scale deployment of the proposed model has the potential to revolutionize mobility assessment in clinical and remote monitoring scenarios. By enabling cost-effective and privacy-compliant evaluations using low-resolution thermal cameras, this methodology could facilitate widespread adoption in healthcare facilities and wearable devices. Future research will explore integrating additional clinical tests, extending applications to fall risk assessment and early detection of mobility disorders, and adapting the model for edge computing platforms to enhance real-time monitoring capabilities.

\section*{Acknowledgment}

We extend our sincere gratitude to the participants in our experiments for their invaluable contribution to the data collection. We thank StreamTeck Corporation for providing the equipment that supported the experiments.

\bibliographystyle{IEEEbib}
\bibliography{strings,refs}
\end{document}